\documentclass[11pt]{article}

\usepackage[a4paper,margin=1.1in,headheight=14pt]{geometry}
\usepackage[T1]{fontenc}
\usepackage[utf8]{inputenc}
\usepackage[sc]{mathpazo}
\usepackage[scaled=0.88]{beramono}
\usepackage{microtype}

\usepackage{amsmath,amssymb,amsthm,bbm}
\usepackage{mathtools}
\usepackage{algorithm}
\usepackage{algpseudocode}

\usepackage{graphicx}
\usepackage{booktabs}
\usepackage{array}
\usepackage{multirow}
\usepackage{subcaption}
\usepackage{float}
\usepackage{placeins}
\usepackage{caption}

\usepackage{enumitem}
\usepackage{xcolor}

\usepackage[numbers,sort&compress]{natbib}
\usepackage[hidelinks,breaklinks=true]{hyperref}
\usepackage[nameinlink,capitalize,noabbrev]{cleveref}
\usepackage{url}

\graphicspath{{figures/}}
\setlist[itemize]{leftmargin=1.35em, itemsep=0.18em, topsep=0.2em}
\setlist[enumerate]{leftmargin=1.4em, itemsep=0.18em, topsep=0.2em}
\usepackage{titlesec}
\titleformat*{\section}{\large\bfseries}
\titleformat*{\subsection}{\normalsize\bfseries}
\titleformat*{\subsubsection}{\small\bfseries}
\titlespacing*{\section}{0pt}{1.1em plus 0.2em minus 0.1em}{0.45em}
\titlespacing*{\subsection}{0pt}{0.8em plus 0.15em minus 0.1em}{0.35em}

\newcommand{\sloguard}{SLO-Guard\xspace}
\usepackage{xspace}
\newcommand{\goodput}{\operatorname{Goodput}}
\newcommand{\ttftp}{\mathrm{TTFT}_{p99}}
\newcommand{\itlp}{\mathrm{ITL}_{p99}}
\DeclareMathOperator*{\argmax}{arg\,max}

\newcommand{\ind}{\mathbbm{1}}

\title{\vspace{-1.5em}\Large\bfseries SLO-Guard: Crash-Aware, Budget-Consistent Autotuning\\for SLO-Constrained LLM Serving\vspace{-0.4em}}
\author{%
  \normalsize Christian Lysenst\o{}en\\[0.15em]
  \small Inland Norway University of Applied Sciences\\[0.05em]
  \small Visiting Student, University of California, Berkeley\\[0.15em]
  \small \texttt{christian@lysenstoen.net} \quad \texttt{github.com/Chrislysen}%
}
\date{\normalsize April 2026}

\begin{document}
\maketitle
\thispagestyle{empty}

\begin{abstract}
\noindent
Serving large language models (LLMs) under latency service-level objectives (SLOs) is a configuration-heavy systems problem with an unusually failure-prone search space: many plausible configurations crash outright or miss user-visible latency targets, and standard black-box optimizers treat these failures as wasted trials. We present \sloguard{}, a crash-aware autotuner for vLLM serving that treats crashes as first-class observations. \sloguard{} combines a feasible-first \emph{Thermal Budget Annealing} (TBA) exploration phase with a warm-started \emph{Tree-structured Parzen Estimator} (TPE) exploitation phase; the handoff replays all exploration history, including crashes encoded as extreme constraint violations. We additionally contribute a configuration-repair pass, a GPU-aware KV-cache memory guard, and a four-category crash taxonomy. On \texttt{Qwen2-1.5B} served with vLLM 0.19 on an NVIDIA A100~40\,GB, and across a pre-specified five-seed study, both \sloguard{} and uniform random search attain $75/75$ feasibility with zero crashes, and are statistically indistinguishable on best-achieved latency (Mann--Whitney two-sided $p\!=\!0.84$). \sloguard{}'s advantage lies in \emph{budget consistency}: it allocates significantly more trials to the fast-serving regime ($10.20{\pm}1.10$ vs.~$7.40{\pm}2.51$ out of 15; one-sided $p\!=\!0.014$) and maintains higher post-handoff consistency ($0.876{\pm}0.123$ vs.~$0.539{\pm}0.224$; $p\!=\!0.010$). Under concurrent load, \sloguard{}'s cross-seed standard deviation on best latency is $4.4{\times}$ tighter than random search's ($2.26$\,ms vs.~$10.00$\,ms). A harness-replication analysis shows that the consistency findings survive an independent sequential-dispatch measurement condition. We present \sloguard{} as a method paper with narrow but replicated empirical validation rather than a general solution to LLM serving autotuning: the central claim is not that \sloguard{} finds a better final configuration, but that it spends a fixed tuning budget more predictably once the fast regime has been found.
\end{abstract}

\vspace{0.3em}

% =========================================================================
\section{Introduction}
\label{sec:intro}

Serving LLMs under latency service-level objectives is now a systems problem as much as a modeling problem. Once a model is chosen, engineers must still select runtime parameters---batch size, token budget, memory utilization, execution mode, prefill strategy, and prefix caching---that interact in non-obvious ways with user-visible latency, memory pressure, and failure risk. A configuration that looks attractive on aggregate throughput may still be unusable if it violates time-to-first-token (TTFT) or inter-token latency (ITL) targets, or if it crashes the serving engine under realistic concurrent request mixes.

Three properties of LLM-serving tuning differentiate it from classical hyperparameter optimization:

\begin{itemize}
  \item \textbf{Failure-prone.} Candidate configurations may crash at engine start, during single-request preflight, or only under realistic concurrent load. These are not observation noise; they are structural holes in the feasible set.
  \item \textbf{Conditional.} Several knobs are meaningful only when others take specific values (e.g.\ \texttt{enable\_chunked\_prefill} is relevant only when \texttt{enforce\_eager}{=}\texttt{false} in vLLM~0.19). A naive Cartesian search space misstates the problem.
  \item \textbf{Small-budget.} Each trial starts a serving engine, warms it up, measures latency under load, and tears it down. In practice, tens of trials is a realistic budget; thousands is not.
\end{itemize}

\paragraph{Research question.} This paper asks a narrow but practically important question: \emph{when many vLLM configurations are crash-prone or unstable, can a crash-aware optimizer use a limited tuning budget more effectively than crash-agnostic baselines?} We are not looking for asymptotic optimality, but for robust behavior under small budgets with mixed discrete/continuous, conditional search spaces and heavy-tailed failure events.

\paragraph{Approach.} We present \sloguard{}, a two-phase crash-aware autotuner. The method extends the author's earlier Thermal Budget Annealing (TBA) work on constrained ML deployment~\citep{Lysenstoen2026TBA} to the LLM inference setting:

\begin{enumerate}
  \item \textbf{TBA-explore}: a feasible-first, crash-aware annealing phase that searches for viable regions and records unstable subspaces.
  \item \textbf{TPE-exploit}: a warm-started Optuna TPE phase that exploits history gathered during exploration, including crashes represented as extreme constraint violations.
\end{enumerate}

A configuration-repair pass and a GPU-aware KV-cache guard bridge the gap between clean mathematical problem statements and the messy reality of a production-style serving API. Both pieces are deliberately boring engineering rather than learned components; their job is to remove implementation-level impossibilities so the optimizer can spend its limited budget on informative decisions.

\paragraph{Key finding.} Our primary empirical result is not a peak-latency win. Under the corrected concurrent harness, both \sloguard{} and random search are 75/75 feasible with zero crashes and are statistically tied on best-achieved latency. What differs is \emph{how the 15-trial budget is spent}. \sloguard{} allocates more trials to the fast-serving regime, stays there more reliably after entering it, and yields substantially tighter cross-seed variance on the best-latency statistic. We therefore frame the contribution around \emph{budget consistency} rather than peak-value superiority.

\paragraph{Contributions.} The paper makes four contributions:
\begin{enumerate}
  \item A crash-aware formulation of SLO-constrained serving autotuning, including the repair map and GPU-aware feasibility guard that make the formulation implementable (\cref{sec:method}).
  \item A simple two-phase optimizer (TBA-explore $\to$ warm-started TPE-exploit) that encodes crashes as extreme constraint violations and replays them through the TPE bad-density model (\cref{sec:method}).
  \item A pre-specified, multi-seed empirical study with non-parametric testing and explicit effect-size reporting, arguing that \sloguard{}'s value lies in consistent budget allocation rather than in final-latency gains (\cref{sec:results}).
  \item A harness-replication analysis showing that the consistency result survives correction of the load-dispatch pipeline from sequential to truly concurrent request dispatch (\cref{sec:harness}).
\end{enumerate}

We position the work as a case study in crash-aware serving autotuning. The method contribution is real, the implementation and data are public, the originally-flagged measurement defect is corrected and replicated, and the conclusions are deliberately kept at the level the evidence supports.

% =========================================================================
\section{Background and Related Work}
\label{sec:related}

\subsection{vLLM, serving knobs, and search-space structure}

vLLM is a widely used LLM serving engine built around PagedAttention-style KV-cache management~\citep{Kwon2023PagedAttention}. The configuration space studied here contains eight knobs: \texttt{quantization}, \texttt{max\_num\_seqs}, \texttt{max\_num\_batched\_tokens}, \texttt{gpu\_memory\_utilization}, \texttt{max\_model\_len}, \texttt{enforce\_eager}, \texttt{enable\_chunked\_prefill}, and \texttt{enable\_prefix\_caching}. Together they span batching, memory pressure, context length, execution mode, and prompt reuse.

Several of these knobs interact structurally in vLLM~0.19: \texttt{enable\_chunked\_prefill} is only meaningful when \texttt{enforce\_eager}{=}\texttt{false}; \texttt{max\_num\_batched\_tokens} must be at least both \texttt{max\_num\_seqs} and \texttt{max\_model\_len}; and the product \texttt{max\_num\_seqs}\,$\times$\,\texttt{max\_model\_len} must fit the available KV-cache budget on the specific GPU. This makes the tuning problem mixed-type, conditional, and hardware-aware.

\subsection{SLOs and goodput}

Raw throughput is insufficient as a standalone metric for user-facing LLM serving because users experience latency, not aggregate token counts~\citep{Wang2024SLO}. Following this line of work we optimize \emph{goodput}: throughput that remains within the latency envelope. Formalism follows in \cref{sec:method}. Our TTFT/ITL thresholds parallel the inference-benchmark conventions of MLPerf~\citep{Reddi2020MLPerf} and the metric set used by NVIDIA's GenAI-Perf measurement tool~\citep{NvidiaGenAIPerf}.

\subsection{Model-serving autotuning}

Auto-configuration for model serving predates the current LLM wave. Morphling~\citep{Wang2021Morphling} uses meta-learning to transfer cloud-native serving configurations across deployments. Managed tuning services such as AWS SageMaker Automatic Model Tuning~\citep{SageMakerAMT} and Google Vizier~\citep{Golovin2017Vizier} are mature examples of black-box optimization as a service. These systems are important context but do not specifically address the question of how to use crash signals when the serving configurations themselves are unstable.

A separate line of systems work focuses on building serving engines and scaling mechanisms rather than autotuning their configurations: Orca~\citep{Yu2022Orca} introduces continuous batching for transformer inference; Alpa~\citep{Zheng2022Alpa} automates inter- and intra-operator parallelism; FlexGen~\citep{Sheng2023FlexGen} demonstrates high-throughput offload-based generative inference; FlashAttention~\citep{Dao2022FlashAttention} shows that kernel-level execution choices can dominate runtime behavior. LLM-specific SLO-directed serving systems include DistServe~\citep{Zhong2024DistServe} and SARATHI-Serve~\citep{Agrawal2024Sarathi}. \sloguard{} is complementary to these: it does not propose a new serving engine, but tunes the knobs that such engines expose.

\subsection{Constrained and crash-aware black-box optimization}

Constrained Bayesian optimization is the closest methodological ancestor. Gelbart et al.~\citep{Gelbart2014BO} treat unknown constraints explicitly within BO; Letham et al.~\citep{Letham2019Constrained} extend the setting to noisy constrained experiments. SMAC~\citep{HutterHoosLeytonBrown2011SMAC} introduced random-forest-based sequential model-based algorithm configuration for noisy evaluations. TPE~\citep{Bergstra2011TPE}, especially as exposed through Optuna~\citep{Akiba2019Optuna}, provides a practical density-ratio view of exploitation in mixed spaces. Under small budgets with heavy-tailed failures and conditional structure, however, the overhead of fitting global surrogates can be prohibitive relative to the measurement cost of each trial.

More directly related is crash-aware or crash-constrained BO. Stenger et al.~\citep{Stenger2024LocalBO} study controller tuning with crash constraints, arguing that crashes should \emph{influence} subsequent search rather than simply terminate trials. \sloguard{} follows the same philosophy but targets vLLM serving specifically and adopts a pragmatic hybrid: simulated-annealing-style feasibility discovery followed by warm-started TPE.

To our knowledge, crash-aware autotuning \emph{specifically} for LLM serving configurations under explicit SLO constraints is not yet well-covered. Adjacent threads exist---serving engines, general serving autotuners, goodput metrics, constrained BO---but \sloguard{} sits at the intersection of all four.

% =========================================================================
\section{Problem Formulation and Method}
\label{sec:method}

\subsection{Feasibility and goodput}

Let $\theta\in\Theta$ denote a (repaired) vLLM configuration and let a trial issue $n$ requests. Define the feasible set
\begin{equation}
\mathcal{F} \;=\; \Bigl\{\, \theta\in\Theta \;:\;
\neg\,\mathrm{crash}(\theta),\;
\ttftp(\theta)\le \tau_{\mathrm{TTFT}},\;
\itlp(\theta)\le \tau_{\mathrm{ITL}},\;
M(\theta)\le \tau_{M} \,\Bigr\},
\label{eq:feasibility}
\end{equation}
where $\tau_{\mathrm{TTFT}}$, $\tau_{\mathrm{ITL}}$, and $\tau_{M}$ are user-specified SLO thresholds for tail time-to-first-token, inter-token latency, and GPU memory. Note that $\neg\,\mathrm{crash}(\theta)$ is a first-class membership condition: a crashing configuration is excluded from $\mathcal{F}$ even if its partial metrics would otherwise satisfy the remaining thresholds.

For a trial with batch wall-clock duration $T(\theta)$ and per-request output token counts $u_1,\dots,u_n$, goodput is
\begin{equation}
\goodput(\theta) \;=\; \frac{1}{T(\theta)}\,\sum_{i=1}^{n}\ind\!\bigl\{\,r_i(\theta)\text{ satisfies the SLO}\,\bigr\}\cdot u_i,
\label{eq:goodput}
\end{equation}
with $\goodput(\theta)=0$ assigned to crashed trials. Under concurrent dispatch, $T(\theta)$ is the elapsed batch wall-clock---the interval from first-issued to last-completed request---and \emph{not} the sum of per-request latencies. This distinction becomes material in \cref{sec:harness}.

The optimization problem is
\begin{equation}
\theta^\star \;\in\; \argmax_{\theta\in\Theta}\;\goodput(\theta)
\quad\text{subject to}\quad \theta\in\mathcal{F}.
\label{eq:problem}
\end{equation}
Crashes are modeled as extreme constraint violations in the implementation: a crashed trial is assigned a large positive violation score and utility $-\infty$ for the purpose of TPE warm-starting.

\subsection{Repair map and GPU-aware memory guard}

Before execution, every proposed $\theta$ passes through a repair map $R:\Theta\to\Theta$ that resolves known vLLM incompatibilities and applies a GPU-aware memory guard. For GPU memory utilization $u$, available VRAM $V$, estimated model footprint $F$, per-token KV-cache cost $\kappa$, and safety margin $\alpha\in(0,1]$, the estimated KV token budget is
\begin{equation}
K_{\max}(u) \;=\; \left\lfloor \alpha\cdot\frac{u\,V - F}{\kappa} \right\rfloor.
\label{eq:kvguard}
\end{equation}
If \texttt{max\_num\_seqs}\,$\times$\,\texttt{max\_model\_len}\,$>K_{\max}(u)$, the map first reduces \texttt{max\_model\_len} and then \texttt{max\_num\_seqs} to restore the inequality, preferring to preserve batch parallelism. $V$ is auto-detected via \texttt{torch.cuda.get\_device\_properties}; $F$ and $\kappa$ are loaded from a per-model registry with a fallback probe of the Hugging Face config for models not in the registry. The intent is not to make all configurations good, but to remove API-level impossibilities that would otherwise masquerade as optimizer failure.

\subsection{TBA-explore phase}

TBA-explore is a feasible-first annealing procedure over the repaired space. The first $n_{\mathrm{init}}$ trials are sampled uniformly at random; subsequent trials are proposed by combining structural mutations (categorical flips, e.g.\ \texttt{enforce\_eager}) with local numeric perturbations, with the structural-move probability annealed over time. Before the first feasible point is found, the optimizer minimizes total constraint violation; after the first feasible point, it switches to goodput optimization within~$\mathcal{F}$. Crashes and infeasible trials are not discarded: they update a per-category bad-region tracker that discourages repeated sampling of structurally-similar configurations.

\begin{algorithm}[H]
\caption{TBA-explore: feasible-first annealing with crash memory}
\label{alg:tba}
\begin{algorithmic}[1]
\Require space $\Theta$, budget $B$, SLO thresholds $C$, repair map $R$
\State Initialize temperature $T$, history $\mathcal{H}\leftarrow\emptyset$, current state $\theta_\star\leftarrow\textsc{None}$
\For{$t = 1,\dots,B$}
  \If{$t\le n_{\mathrm{init}}$ or $\theta_\star=\textsc{None}$}
    \State Sample $\theta_t$ uniformly at random from $\Theta$
  \Else
    \State Propose $\theta_t$ as a temperature-$T$ neighbor of $\theta_\star$
  \EndIf
  \State $\theta_t \leftarrow R(\theta_t)$ \Comment{apply repair and KV-cache guard \eqref{eq:kvguard}}
  \State Execute benchmark; observe crash flag $c_t$, metrics $m_t$
  \State $\mathcal{H}\leftarrow\mathcal{H}\cup\{(\theta_t,c_t,m_t)\}$
  \If{$c_t=1$}
    \State Record crash category $\kappa_t$ in bad-region tracker; decay $T$ as non-improving step; \textbf{continue}
  \EndIf
  \State Compute total constraint violation $v_t$ from $m_t$ and $C$
  \If{$\theta_\star=\textsc{None}$}
    \If{$v_t=0$} \State $\theta_\star\leftarrow\theta_t$ \Comment{first feasible point}
    \Else \State accept $\theta_t$ with probability $\exp(-\Delta v / T)$; update $\theta_\star$
    \EndIf
  \Else
    \State accept $\theta_t$ with probability $\min\{1,\exp(\Delta \goodput / T)\}$; update $\theta_\star$
  \EndIf
  \State Update temperature $T$
  \If{handoff condition (\cref{alg:handoff}) satisfied} \State \textbf{break} \EndIf
\EndFor
\State \Return $\mathcal{H}$
\end{algorithmic}
\end{algorithm}

\subsection{Warm-started TPE exploitation}

After sufficient feasible and bad-region history accumulates, \sloguard{} hands off to Optuna's TPE sampler with zero startup trials. All TBA history, including crashes (encoded as large positive constraint violations), is replayed into the Optuna study before the exploit phase begins. TPE then maintains two kernel-density estimates over the promising and non-promising regions of $\Theta$, denoted $\ell(\theta)$ and $g(\theta)$ respectively, and samples candidates with high ratio
\begin{equation}
\theta_t \;\in\; \argmax_{\theta\in\Theta}\;\frac{\ell(\theta)}{g(\theta)}.
\label{eq:tpe}
\end{equation}
The warm start matters: a cold-start TPE with only a handful of trials has almost no density signal at this budget, whereas warm-started TPE immediately sees both the feasible region and the crash boundary.

\begin{algorithm}[H]
\caption{Phase transition from TBA-explore to warm-started TPE-exploit}
\label{alg:handoff}
\begin{algorithmic}[1]
\Require TBA history $\mathcal{H}$, budget $B$
\State $t_{\min}\leftarrow\max(3,\lfloor B/5\rfloor)$; $\;t_{\max}\leftarrow\max(5,\lfloor 0.4B\rfloor)$
\If{$|\mathcal{H}|<t_{\min}$} \State \Return \textsc{stay-in-tba}
\ElsIf{$|\mathcal{H}|\ge t_{\max}$} \State \textbf{goto} \textsc{handoff}
\Else
  \State $n_f\leftarrow\#\{$feasible trials in $\mathcal{H}\}$; $\;n_b\leftarrow\#\{$crashes or SLO-violating trials in $\mathcal{H}\}$
  \If{$n_f\ge n_f^{\min}$ and $n_b\ge n_b^{\min}$}
    \State \textbf{goto} \textsc{handoff}
  \Else
    \State \Return \textsc{stay-in-tba}
  \EndIf
\EndIf
\Statex \textsc{handoff:}
\State Instantiate Optuna study with constrained TPE sampler and zero startup trials
\ForAll{$(\theta_i,c_i,m_i)\in\mathcal{H}$}
  \State Inject as completed trial; crashes recorded with large violation score
\EndFor
\State \Return \textsc{run-tpe-for-remainder}
\end{algorithmic}
\end{algorithm}

For the 15-trial budget used throughout this paper, the handoff empirically occurs at trial~7 in all reported runs.

\subsection{Crash taxonomy}
\label{sec:crash-taxonomy}

Each trial is classified into one of four categories according to where in the pipeline the outcome is determined:
\begin{itemize}
  \item \textbf{Healthy}: engine started, preflight completed, all benchmark requests returned; goodput is computable.
  \item \textbf{Startup failure}: engine never reached \texttt{/health} within the startup timeout (typically KV-cache warmup OOM or CUDA init failure).
  \item \textbf{Preflight failure}: engine healthy but cannot serve a single-token completion (typically an invalid flag combination such as \texttt{enforce\_eager}{=}\texttt{true} with \texttt{enable\_chunked\_prefill}{=}\texttt{true} in vLLM~0.19).
  \item \textbf{Runtime failure}: preflight succeeded but the benchmark phase returned 500-class errors or connection resets under load.
\end{itemize}
These categories are not cosmetic. Startup failures are geometry/memory problems in $\mathcal{F}$; preflight failures are flag-combination problems; runtime failures indicate stability under load. The bad-region tracker is therefore conditioned on category, which makes crash observations generalize across nearby configurations rather than attach to a single failed point.

% =========================================================================
\section{Experimental Setup}
\label{sec:setup}

\paragraph{Hardware and software.} All results use Google Colab A100~40\,GB, model \texttt{Qwen/Qwen2-1.5B}, vLLM~0.19, Optuna for TPE, and BoTorch for the (unused-in-this-paper) constrained BO baseline. Serving processes are launched with \texttt{preexec\_fn=os.setsid} so they can be hard-killed at the process-group level between trials.

\paragraph{Workload and SLOs.} Each trial issues five requests at a nominal target rate of $1$~req/s with a fixed-template prompt and a 100-token output cap. Under the concurrent harness---our primary condition---the five requests are dispatched with overlap subject to a bounded-concurrency semaphore, and $T(\theta)$ in \cref{eq:goodput} is measured as batch wall-clock. SLO thresholds are $\ttftp\le 500$\,ms, $\itlp\le 100$\,ms, with memory feasibility enforced by \cref{eq:kvguard}.

\paragraph{Optimizers compared.} The repository implements uniform random search, cold-start Optuna TPE, TBA-only, \sloguard{} (TBA-TPE hybrid), and a GP-based constrained BO baseline via BoTorch. The primary head-to-head reported at the full multi-seed scale is \textbf{Random vs.\ \sloguard{}}. We use uniform random search as the primary baseline because, at 15 trials with a discrete-dominated conditional space, random is the tougher reference; more sophisticated surrogates typically need more evaluations to separate from uniform sampling in this regime~\citep{HutterHoosLeytonBrown2011SMAC}. The additional baselines are discussed as future work (\cref{sec:limitations}).

\paragraph{Multi-seed protocol.} The main study is $5\text{ seeds}\times 2\text{ optimizers}\times 15\text{ trials}=150$ trials per measurement condition, with seeds $\{42,142,242,342,442\}$. The runner alternates optimizers within seed so that partial completion remains balanced across Colab disconnections, and persists each trial result with \texttt{flush()}+\texttt{fsync()} so interrupted runs are recoverable via replay.

\paragraph{Metrics.} We report three per-seed summary statistics:
\begin{enumerate}
  \item \textbf{Fast-cluster trials} (out of 15): number of feasible trials with average request latency below $1000$\,ms. The $1000$\,ms threshold separates the two visibly distinct latency bands in this workload (\cref{sec:bimodal}) and was chosen for that structural reason rather than to maximize the method's apparent advantage.
  \item \textbf{Post-hit consistency}: let $t^\star$ be the first trial with average latency below $1000$\,ms; consistency is $\#\{t>t^\star:\text{trial }t\text{ is fast-cluster}\}/\#\{t>t^\star\}$. It answers: \emph{after the fast regime has been found, what fraction of remaining trials stay there?}
  \item \textbf{Best latency}: $\min_t\text{avg.~latency}$ over feasible trials for the $(\text{optimizer},\text{seed})$ pair.
\end{enumerate}

\paragraph{Statistical methodology.} Pairwise comparisons use Mann--Whitney $U$ tests on per-seed summaries. For fast-cluster count and post-hit consistency, the alternative is one-sided (\sloguard{} $>$ Random), matching the directional hypothesis that a crash-aware optimizer should allocate budget to the right regime more reliably than blind sampling. For best latency, no directional claim is made, and the test is two-sided. We report raw $p$-values and also apply Holm--Bonferroni correction across the three-metric family; both are shown. Effect sizes are reported as cross-seed variance ratios $\sigma^{2}_{\text{Random}}/\sigma^{2}_{\text{\sloguard{}}}$ and per-seed win counts. With five seeds, the study is intentionally sized to detect only large and consistent effects, and every statistical claim is reported in that spirit.

% =========================================================================
\section{Results}
\label{sec:results}

\subsection{Primary multi-seed result under concurrent dispatch}
\label{sec:results-primary}

\cref{tab:concurrent} reports the head-to-head under the concurrent harness. Both optimizers reach the feasibility ceiling, and best-achieved latency is statistically tied. The structural differences appear on the budget-allocation metrics.

\begin{table}[H]
\centering
\small
\caption{Primary concurrent-harness result. Values are mean $\pm$ standard deviation across five seeds. $p$-values from Mann--Whitney $U$ tests: one-sided (\sloguard{} $>$ Random) for the consistency metrics, two-sided for best latency. Holm--Bonferroni-adjusted $p$-values across the three-metric family are $p_{\text{adj}}=0.028,\,0.030,\,1.00$ respectively.}
\label{tab:concurrent}
\begin{tabular}{lccc}
\toprule
Metric & Random ($n{=}5$) & \sloguard{} ($n{=}5$) & Mann--Whitney $p$ \\
\midrule
Fast-cluster trials / 15 \ (higher is better) & $7.40 \pm 2.51$ & $\mathbf{10.20 \pm 1.10}$ & $\mathbf{0.014}$ \\
Post-hit consistency \ (higher is better)      & $0.539 \pm 0.224$ & $\mathbf{0.876 \pm 0.123}$ & $\mathbf{0.010}$ \\
Best latency, ms \ (lower is better)           & $470.5 \pm 10.00$ & $465.7 \pm 2.26$          & $0.84$ (tied) \\
\midrule
Feasibility / 75 & $75/75$ & $75/75$ & --- \\
Crashes          & $0$     & $0$     & --- \\
\bottomrule
\end{tabular}
\end{table}

\sloguard{} wins on fast-cluster count in 4 of 5 seeds with 1 tie: Random $\{9,8,3,9,8\}$ vs.\ \sloguard{} $\{9,9,11,11,11\}$ for seeds $\{42,142,242,342,442\}$. The most dramatic difference is on seed 242, where Random places only 3 of 15 trials in the fast regime while \sloguard{} places 11. This asymmetry motivates reporting \emph{worst-case} budget usage alongside mean behavior, not only peak performance.

\subsection{How the budget-allocation advantage is spent}
\label{sec:results-trajectory}

\begin{figure}[H]
\centering
\includegraphics[width=0.94\textwidth]{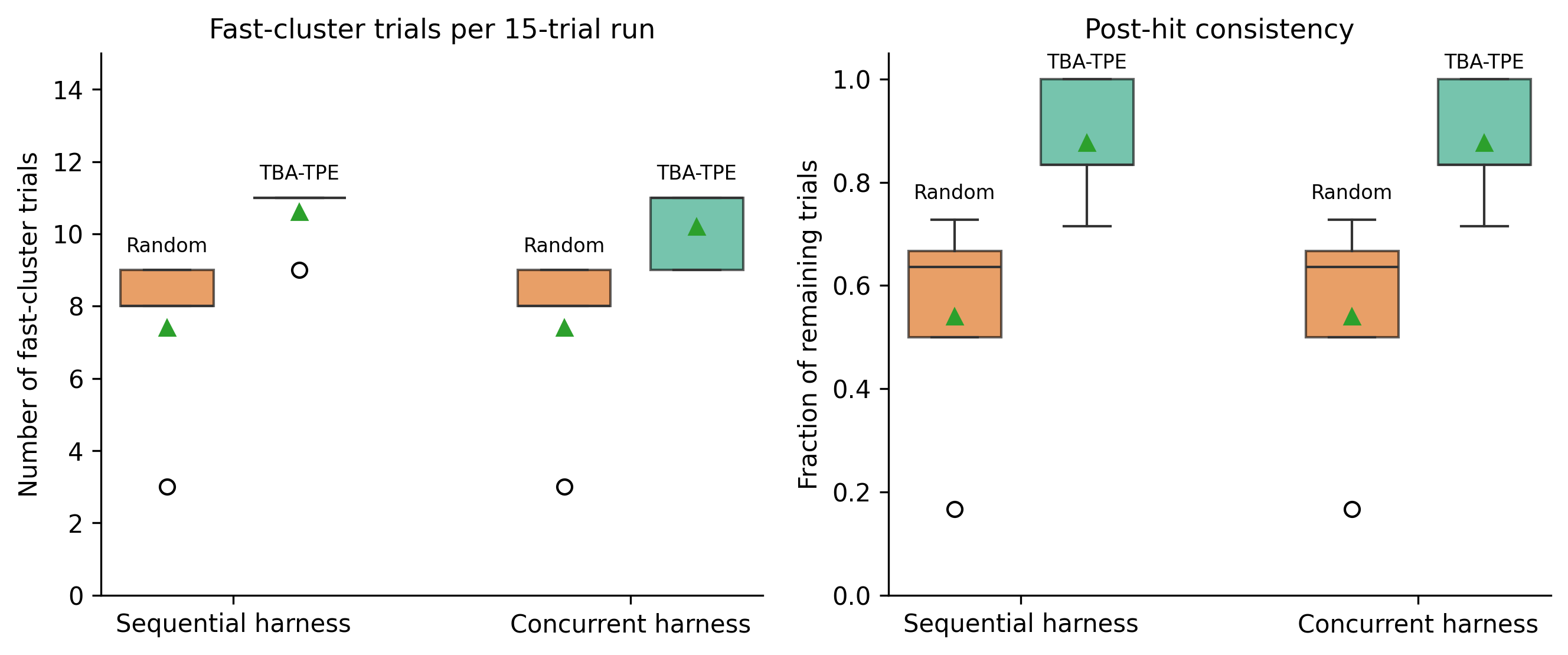}
\caption{Primary consistency figures (concurrent harness). \emph{Left:} per-seed count of fast-cluster trials (avg.~latency $<1000$\,ms) out of 15. \emph{Right:} post-hit consistency---fraction of trials remaining in the fast cluster after the first fast trial. \sloguard{} is both \emph{higher} and \emph{less variable} on both metrics.}
\label{fig:consistency}
\end{figure}

\cref{fig:consistency} visualizes the two primary statistics. The left panel shows per-seed fast-cluster counts; the right panel shows post-hit consistency. Both panels move in the same direction and support the headline framing.

\begin{figure}[H]
\centering
\includegraphics[width=0.94\textwidth]{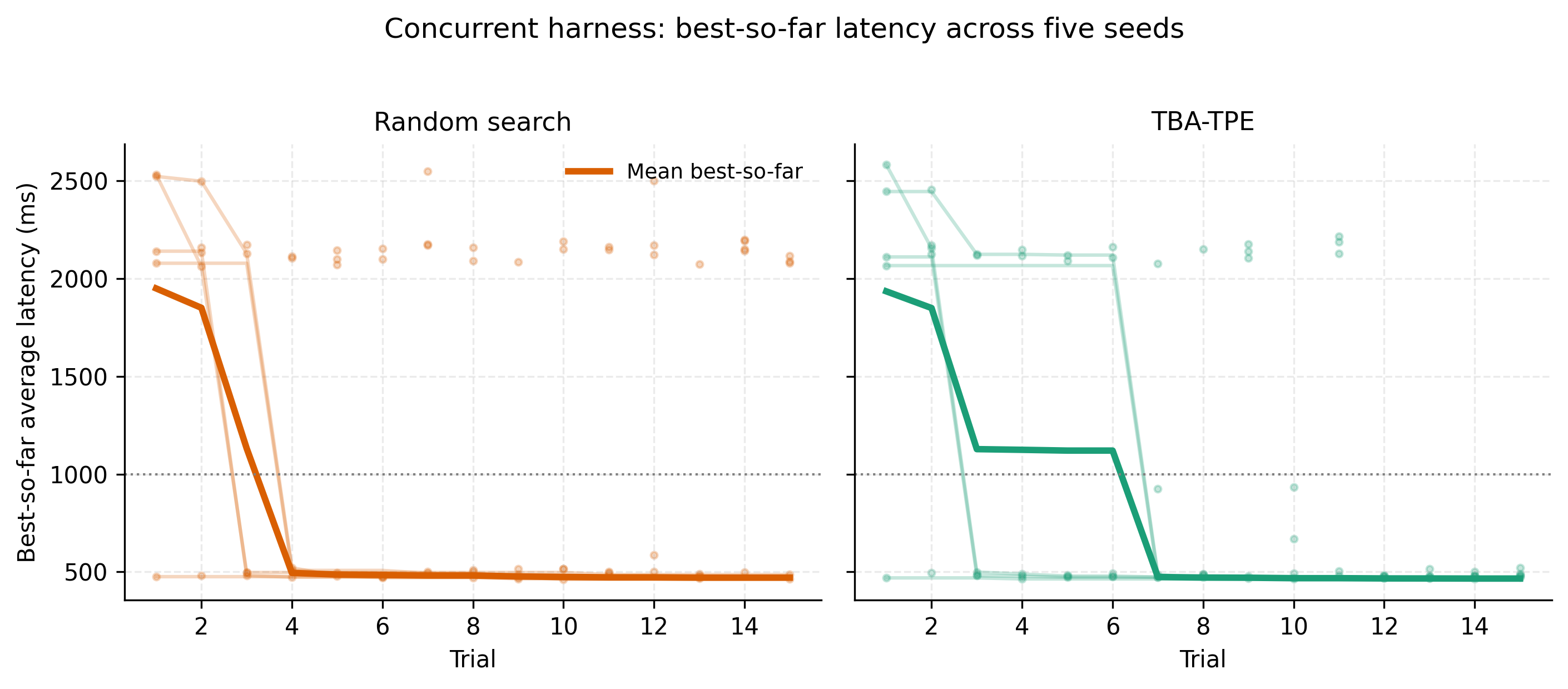}
\caption{Best-so-far average latency over trials for all five seeds under the concurrent harness. \emph{Left:} random search. \emph{Right:} \sloguard{}. The important visual fact is not only where the curves end, but how quickly they stop wandering: random occasionally jumps back to slow configurations after finding fast ones, whereas \sloguard{} settles after handoff and refines within the fast regime.}
\label{fig:convergence}
\end{figure}

\cref{fig:convergence} plots best-so-far latency across all five seeds. Random search sometimes enters the fast regime earlier than \sloguard{}, but several random trajectories continue to sample slower configurations after already finding a fast one. \sloguard{} deliberately spends early trials exploring structure; once its TBA-to-TPE handoff fires, the trajectories flatten and remain inside the fast regime. This is the trajectory-level explanation for the post-hit consistency statistic in \cref{tab:concurrent}.

If one measures only \emph{time to first fast point}, random is competitive and in some seeds better. If one measures \emph{consistency after a good point is found}, \sloguard{} is substantially better. This is why the paper's headline metric is post-hit consistency, not time-to-first-hit.

\subsection{Best-latency spread: tied means, tighter variance}
\label{sec:results-variance}

\begin{figure}[H]
\centering
\includegraphics[width=0.64\textwidth]{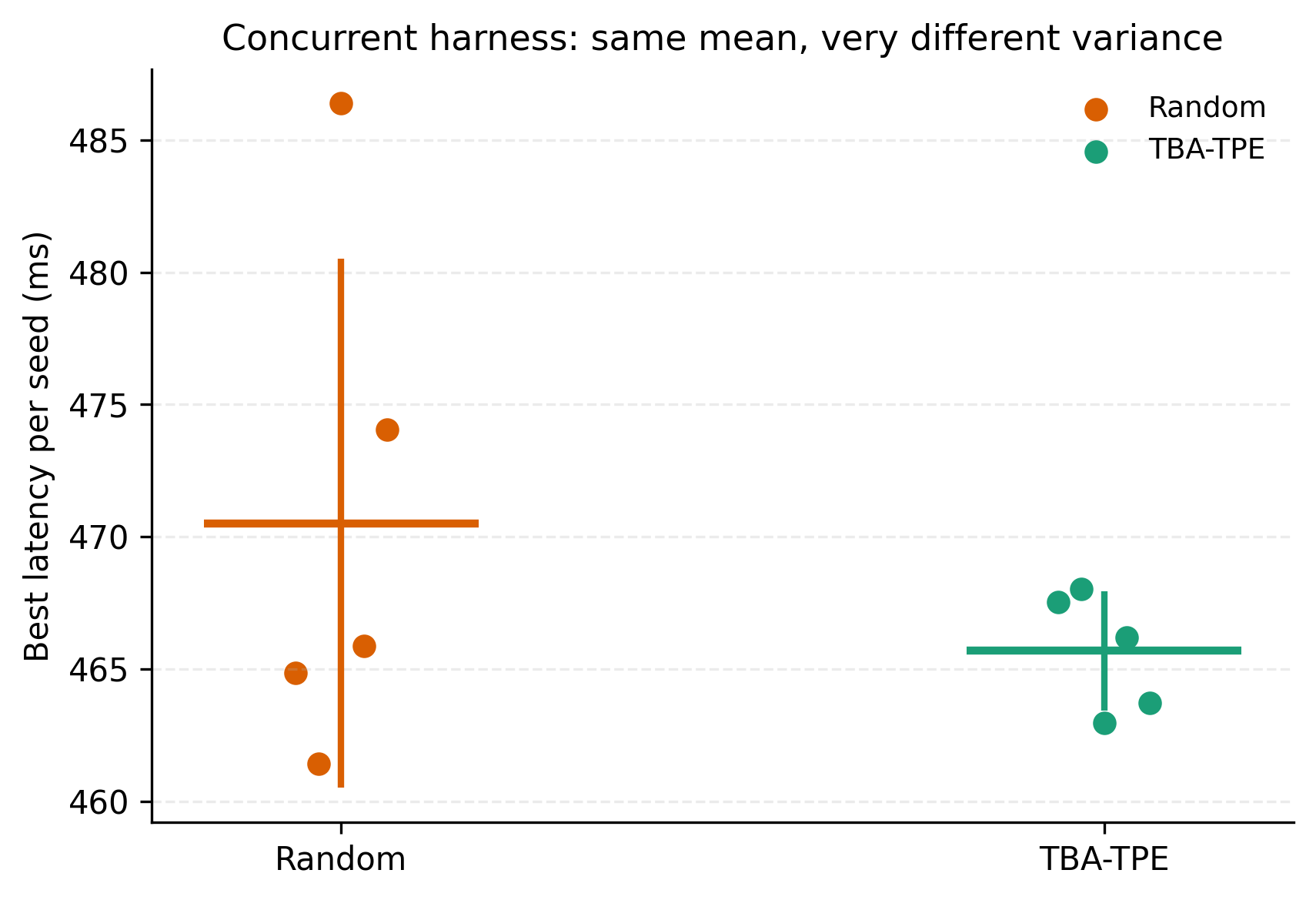}
\caption{Best latency reached by each seed under the concurrent harness. Means are statistically tied (two-sided Mann--Whitney $p=0.84$); the cross-seed spread is not. Random $\sigma\!=\!10.00$\,ms vs.\ \sloguard{} $\sigma\!=\!2.26$\,ms, a ratio of $4.42{\times}$.}
\label{fig:latencyvar}
\end{figure}

\cref{fig:latencyvar} shows per-seed best-achieved latency. Means are statistically indistinguishable, so we do \emph{not} claim a best-latency win. What differs is the cross-seed spread: Random's best-latency standard deviation is $10.00$\,ms and \sloguard{}'s is $2.26$\,ms, a ratio of $4.42{\times}$.

This matters in deployment: an operator does not run a tuning job once. A method that occasionally lands on an excellent point but varies widely across restarts is less attractive than one that reaches essentially the same good point every time. The graph justifies the ``tied means, tighter variance'' wording without over-claiming.

\subsection{Harness replication: sequential vs.\ concurrent dispatch}
\label{sec:harness}

An earlier version of the measurement pipeline serialized requests: successive requests were issued only after the previous one returned, so despite a nominal concurrency setting the actual in-flight count was one. We flag this openly, correct it by replacing the loop with an \texttt{asyncio.Semaphore}-capped concurrent dispatcher, and re-ran the complete optimizer/seed matrix through the corrected harness. Per-trial batch-wall-clock times in the published JSONL confirm the correction: slow-cluster trials with per-request latency $\approx 2500$\,ms now exhibit batch wall-clock $\approx 4100$\,ms (real overlap), not $\approx 12500$\,ms (serialized). Both data sets are retained and published.

\begin{figure}[H]
\centering
\includegraphics[width=0.98\textwidth]{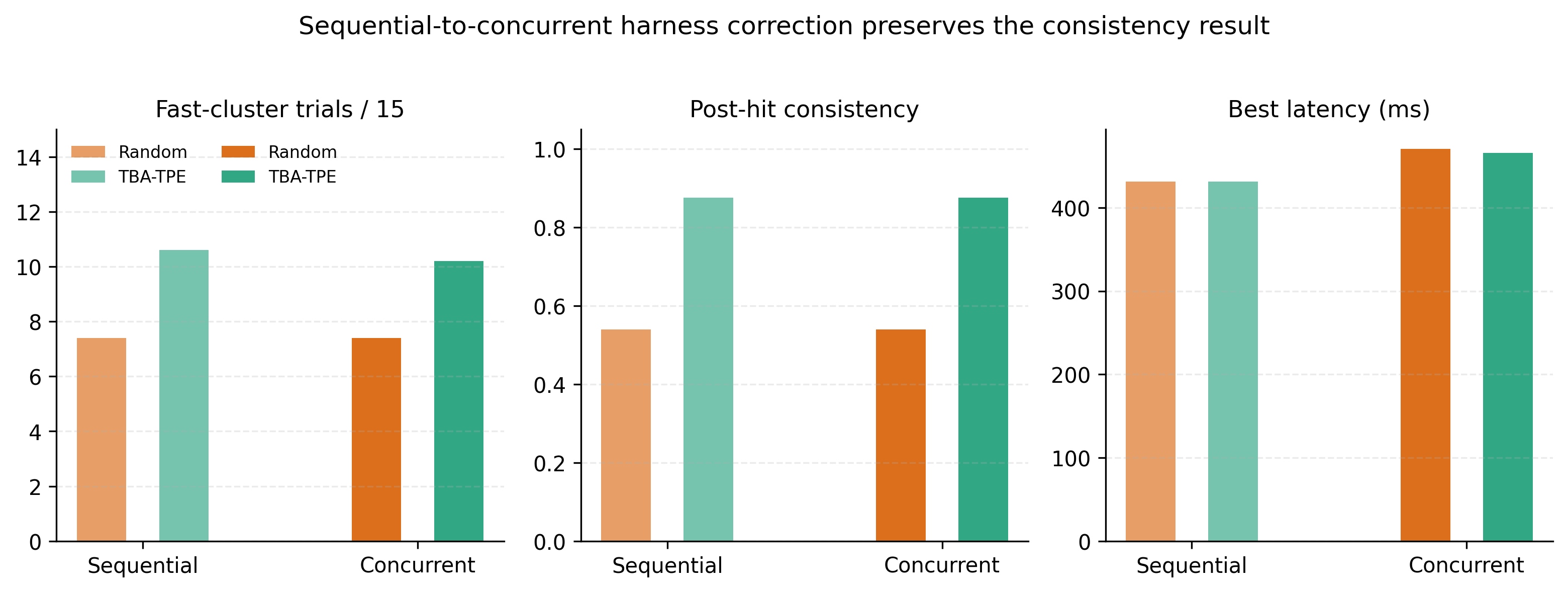}
\caption{Sequential-versus-concurrent harness comparison. Consistency statistics move negligibly under correction; absolute latency rises and its variance becomes more realistic. The central conclusion survives; the corrected harness additionally reveals a variance structure that sequential dispatch compressed away.}
\label{fig:harness}
\end{figure}

\cref{tab:harness} reports all three metrics under both conditions; \cref{fig:harness} plots them side by side.

\begin{table}[H]
\centering
\small
\caption{Harness replication. The consistency conclusions survive the transition from sequential to concurrent measurement. The best-latency variance advantage is visible only under concurrent load.}
\label{tab:harness}
\begin{tabular}{lcccc}
\toprule
 & \multicolumn{2}{c}{Sequential harness} & \multicolumn{2}{c}{Concurrent harness} \\
\cmidrule(lr){2-3}\cmidrule(l){4-5}
Metric & Random & \sloguard{} & Random & \sloguard{} \\
\midrule
Feasibility / 75                          & $75$ & $75$ & $75$ & $75$ \\
Crashes                                   & $0$  & $0$  & $0$  & $0$  \\
Fast-cluster trials / 15                  & $7.40\pm 2.51$ & $10.60\pm 0.89$ & $7.40\pm 2.51$ & $10.20\pm 1.10$ \\
Post-hit consistency                      & $0.539\pm 0.224$ & $0.876\pm 0.123$ & $0.539\pm 0.224$ & $0.876\pm 0.123$ \\
Best latency, ms                          & $431.1\pm 1.74$ & $431.6\pm 1.90$ & $470.5\pm 10.00$ & $465.7\pm 2.26$ \\
\midrule
$p$ (fast-cluster, one-sided)             & \multicolumn{2}{c}{$0.008$} & \multicolumn{2}{c}{$0.014$} \\
$p$ (post-hit, one-sided)                 & \multicolumn{2}{c}{$0.010$} & \multicolumn{2}{c}{$0.010$} \\
$p$ (best latency, two-sided)             & \multicolumn{2}{c}{$0.84$}  & \multicolumn{2}{c}{$0.84$} \\
Variance ratio $\sigma^2_{R}/\sigma^2_{S}$ on best latency & \multicolumn{2}{c}{$0.84{\times}$} & \multicolumn{2}{c}{$\mathbf{4.42{\times}}$} \\
\bottomrule
\end{tabular}
\end{table}

That two independent measurement protocols yield statistically equivalent conclusions on the consistency metrics is direct evidence that those findings are not an artifact of the load-generator implementation. The best-latency variance advantage is specific to the concurrent condition because only under genuine concurrent load does the fast cluster exhibit measurable within-cluster variation (from scheduler jitter and contention inside vLLM).

\subsection{Structural context: the search space is sharply bimodal}
\label{sec:bimodal}

\begin{figure}[H]
\centering
\includegraphics[width=0.88\textwidth]{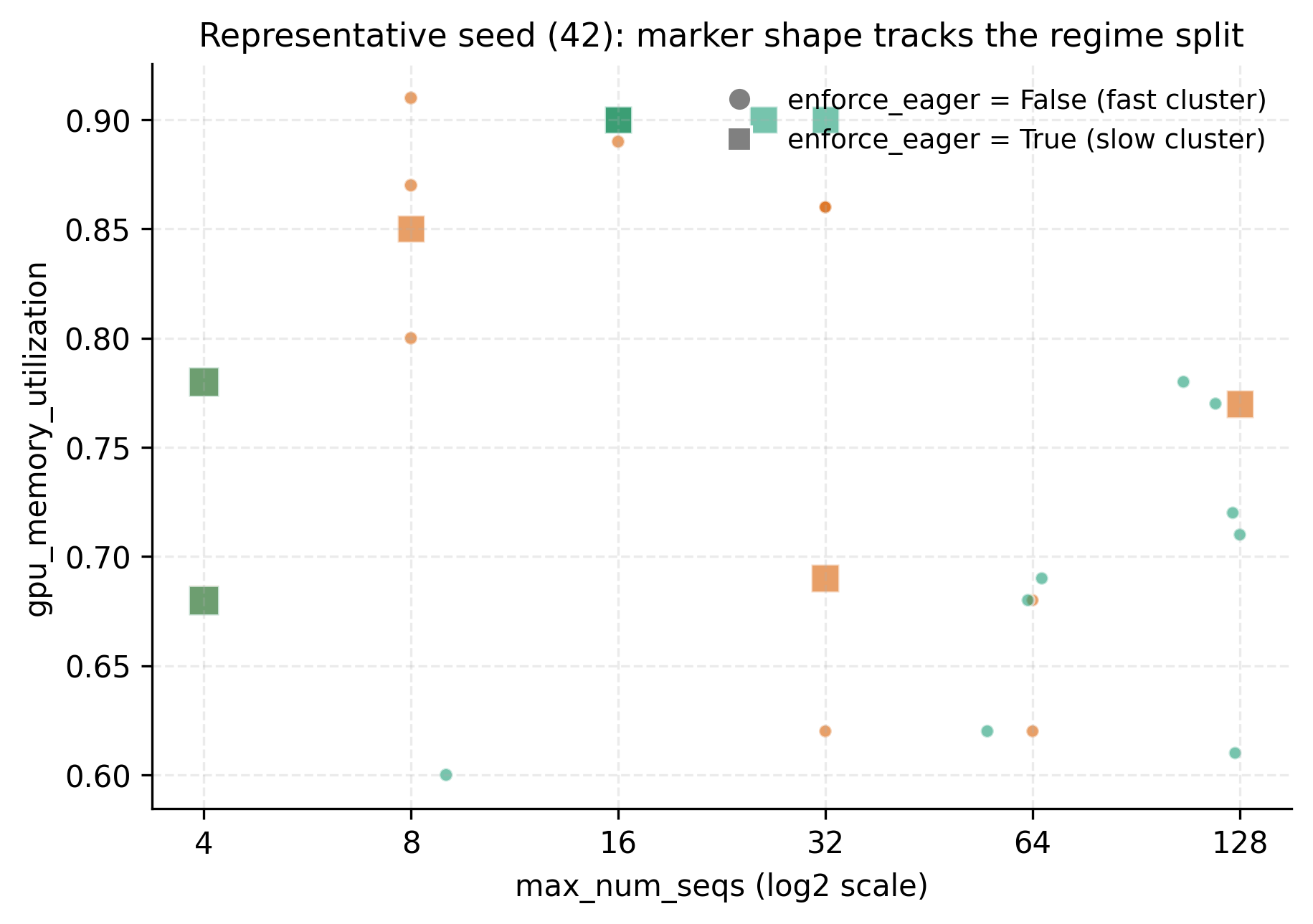}
\caption{Trial-level view of the concurrent-harness search space. The dominant explanatory variable is \texttt{enforce\_eager}: \texttt{true} $\Rightarrow$ slow regime, \texttt{false} $\Rightarrow$ fast regime. This structural observation is why the paper frames its contribution as regime-discovery consistency rather than general many-knob co-optimization.}
\label{fig:bimodal}
\end{figure}

\cref{fig:bimodal} provides a trial-level scatter of the concurrent-harness search space. Feasible points separate into two distinct latency bands. The dominant explanatory variable is a single binary knob: configurations with \texttt{enforce\_eager}{=}\texttt{true} populate the slow regime ($\sim\!2100$\,ms) and those with \texttt{enforce\_eager}{=}\texttt{false} populate the fast regime ($\sim\!470$\,ms). In our sample, this single knob explains over 95\% of latency variance.

This structure keeps the paper honest. A smooth, gently varying landscape would invite broader autotuning claims. A regime-switching landscape dominated by one binary knob reframes the problem: the first-order task is finding the right setting of \texttt{enforce\_eager}; only then does fine-grained knob tuning matter. Crash-aware small-budget search is well-suited to regime discovery, which is why the contribution is still interesting, but the generalization claim must be kept narrow.

\subsection{Representative phase-transition trajectory}
\label{sec:phase}

\begin{figure}[H]
\centering
\includegraphics[width=0.88\textwidth]{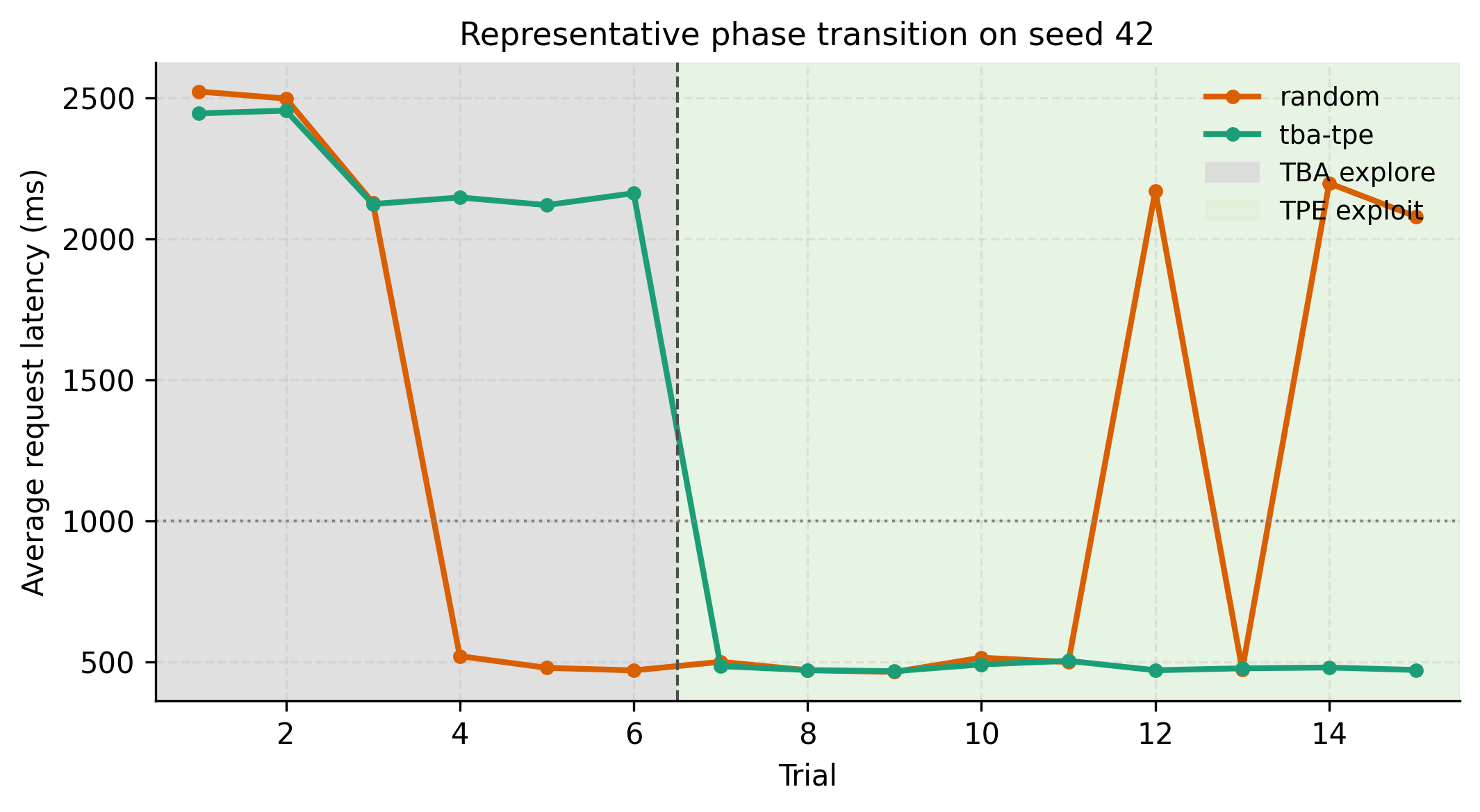}
\caption{Representative seed-42 trajectory for \sloguard{} under the concurrent harness. Trials 1--6: TBA-explore (slow regime). Trial 7: handoff to TPE-exploit. Trials 7--15: fast regime, with variance tightening as TPE concentrates sampling. The handoff corresponds to a real change in search behavior rather than a cosmetic phase label.}
\label{fig:phase}
\end{figure}

\cref{fig:phase} shows a single-seed view of the TBA-to-TPE handoff. Trials 1--6 are TBA-explore, of which all land in the slow cluster; trial 7 marks the handoff to TPE-exploit, which immediately enters the fast cluster and remains there for trials 7--15. The figure is a mechanistic illustration, not a proof; the aggregate argument is \cref{fig:consistency,fig:convergence}.

\subsection{Ablations the evidence suggests}

The current results motivate several natural ablations:
\begin{enumerate}
  \item \textbf{TBA-only vs.\ TBA-TPE.} Does the handoff help, or is pure TBA equally good at this budget?
  \item \textbf{Cold-start TPE vs.\ warm-started TPE.} How much of the gain comes from replaying crash-annotated history into the TPE study, as opposed to using TPE alone?
  \item \textbf{Repair map on/off.} Does the GPU-aware memory guard (\cref{eq:kvguard}) account for the zero-crash rate across 150 concurrent-harness trials?
  \item \textbf{Fix \texttt{enforce\_eager}.} With the dominant binary knob held constant, does the consistency advantage persist as a within-cluster fine-tuning effect, or does it collapse?
\end{enumerate}
The repository implements the other baselines, but the multi-seed ablation study at this scale is left to a follow-up paper.

% =========================================================================
\section{Discussion}
\label{sec:discussion}

The results support a specific and deliberately limited interpretation. \sloguard{} is not winning because it finds a uniquely better final configuration; it is winning because, under a small budget, it allocates more trials to the correct regime and spends fewer late trials wandering back into slow configurations after already finding a fast one. The right verbal summary is therefore \emph{crash-aware two-phase search improves budget consistency under fixed-budget vLLM autotuning}, not \emph{\sloguard{} solves LLM serving autotuning}.

\paragraph{When should crash-awareness help?} The evidence suggests a testable answer: crash-awareness helps most when the search budget is small, the feasible set is ragged, and regime switches exist that a cold-start optimizer can enter but not reliably stay within. In deployments where the feasible set is simply connected and crashes are rare, we expect the advantage to shrink; cold-start BO or even random search should catch up. Testing that regime is future work.

\paragraph{Deployment takeaways.} Two practical lessons are robust across our experiments. First, auto-detect the GPU and apply a memory guard; a hard-coded VRAM assumption is a reliable way to silently inflate crash rates on non-A100 hardware. Second, measure with real concurrent load: the sequential/concurrent comparison in \cref{sec:harness} shows that the fast cluster has $\sim\!40$\,ms of additional per-config variation under concurrent load that is invisible under serialized dispatch. Ranking optimizers on sequentially-measured latency can hide consistency advantages that matter in production.

\paragraph{What the bimodal finding actually shows.} The dominant factor in this evaluation is a single binary switch related to eager execution and CUDA-graph use. That makes the tuning problem simpler than the fully general LLM-serving tuning problem: realistic deployments can exhibit stronger interactions among sequence count, batched-token budget, KV-cache sizing, long-context behavior, and prefill strategy. The present paper should be read as a case study of one crash-aware tuning setting, not a universal statement about LLM autotuning. The correct next step is not ``claim bigger''; it is to check whether the same consistency story holds when the workload structure is richer.

% =========================================================================
\section{Threats to Validity and Limitations}
\label{sec:limitations}

\paragraph{Construct validity.} The study optimizes goodput under TTFT/ITL constraints, following~\citep{Wang2024SLO}. The construct assumes that tail latencies over short benchmark runs generalize to sustained serving behavior. Production serving is bursty; sustained workloads may stress configurations differently.

\paragraph{Internal validity.} Feasibility was observed to be stochastic across repeated Colab runs in early engineering work. The repair map and GPU-aware memory guard appear to have stabilized feasibility at $100\%$ on the reported runs, but the feasibility boundary may still partly depend on environment state not observable from within a Colab notebook.

\paragraph{External validity.} All multi-seed results use one model (\texttt{Qwen/Qwen2-1.5B}) and one GPU (A100 $40$\,GB). Larger models, multi-GPU serving, and consumer GPUs introduce different memory geometries and different dominant knobs. The bimodal \texttt{enforce\_eager} effect may not transfer.

\paragraph{Statistical validity.} Five seeds is deliberately modest. Under non-parametric testing, this budget only detects large, consistent effects. We report effect sizes (variance ratios, per-seed win counts) explicitly and apply Holm--Bonferroni across the three-metric family; both adjusted and raw $p$-values are shown. The replication across two measurement harnesses strengthens the statistical case for the consistency claims.

\paragraph{Workload realism.} The benchmark prompt is a fixed template with a 100-token output cap. Realistic traffic is bursty, varied, and includes multi-turn interactions; the present study does not test those conditions. The corrected harness does introduce genuine concurrent overlap, but not full trace replay.

\paragraph{Baseline coverage.} The repository includes TBA-only, cold-start TPE, and GP-based constrained BO baselines, but the full multi-seed result reported here is \textbf{Random vs.\ \sloguard{}}. That suffices for the narrow claim; it does not establish dominance over all reasonable alternatives.

\paragraph{Absolute tuning-cost objective.} The \sloguard{} implementation includes a utility-style objective that can penalize evaluation cost directly. The main-text experiments retain goodput as the objective for comparability with the earlier pilot; full utility-objective results are left for follow-up work.

\paragraph{Pending citation details.} Two citations to the author's own pending 2026 preprints use placeholder entries; arXiv identifiers will be filled in on public release.

% =========================================================================
\section{Reproducibility}
\label{sec:repro}

Code, raw per-trial JSONL logs under both harness conditions, summary statistics, and figures are public at the project repository~\citep{Lysen2026SLOGuard}. Each trial is logged with the full repaired configuration, the crash category (if any), per-request latencies, batch wall-clock, and aggregate metrics. The multi-seed runner is resumable: it skips completed $(\text{optimizer},\text{seed})$ pairs and, for partially-completed pairs, replays the persisted trials through a fresh optimizer instance so that the optimizer state is consistent with the partial log before resuming. This matters because the TBA-to-TPE handoff depends on history, not merely on trial count.

The concurrent harness records batch wall-clock time per trial, allowing the reader to verify concurrency at the record level (a truly concurrent five-request batch with per-request latency $\approx 2500$\,ms has batch wall-clock $\approx 4000$\,ms, not the $\approx 12500$\,ms that a serialized five-request loop would produce). The statistical analysis is reproduced by \texttt{scripts/compute\_multiseed\_stats.py}.

Commit hashes corresponding to the numbers in this paper:
\begin{itemize}
  \item Sequential-harness data: \texttt{54bddd2}.
  \item Concurrent-harness data: \texttt{f26022f}.
  \item Load-generator concurrency fix: \texttt{d4cbc15}.
  \item Utility-objective implementation: \texttt{68667c9}.
  \item Metrics-collector correction: \texttt{2f94f1a}.
\end{itemize}
To reproduce the concurrent-harness study, clone the repository at commit \texttt{f26022f} and run
\begin{center}\small
\texttt{python scripts/run\_multiseed.py --output-dir results/multiseed\_concurrent/ \textbackslash{}}\\
\texttt{\quad --seeds 42 142 242 342 442 --optimizers random tba-tpe \textbackslash{}}\\
\texttt{\quad --budget 15 --model Qwen/Qwen2-1.5B}
\end{center}
on an A100 $40$\,GB instance.

% =========================================================================
\section{Conclusion and Future Work}
\label{sec:conclusion}

\sloguard{} is a crash-aware two-phase autotuner for vLLM serving: a feasible-first TBA exploration phase that treats crashes as first-class observations, followed by a warm-started TPE exploitation phase that exploits the full exploration history. Across 150 concurrent-harness trials on \texttt{Qwen2-1.5B}/A100, and a second 150-trial sequential-harness replication, the optimizer does not achieve a significantly lower best latency than uniform random search, but it uses a fixed 15-trial budget more consistently: more fast-regime trials, higher post-handoff consistency, and under concurrent load a $4.4{\times}$ tighter cross-seed best-latency spread.

The paper's strongest contribution is not the optimizer in isolation but the combination of the method with a careful, replicated empirical argument about what the method actually improves. The next steps are concrete: complete the baseline matrix (cold-start TPE, TBA-only, constrained BO) at the multi-seed scale, to disentangle the contribution of warm-starting from the contribution of crash-aware exploration per se; test whether the consistency story survives on a second model class (e.g.\ Phi-2 or Llama-3.2-1B) and a second hardware class (e.g.\ L4 or T4); and move from fixed synthetic prompts to realistic traces with burstiness and multi-turn structure. An honest autotuning claim for production serving must ultimately be measured against production-shaped traffic; this paper is the first step toward that measurement.

% =========================================================================
\section*{Acknowledgments}

This work was conducted as a student research project at Inland Norway University of Applied Sciences while the author was on exchange at the University of California, Berkeley. It benefited from the open-source vLLM, Optuna, and BoTorch ecosystems. An external reviewer of an early draft flagged the load-generator concurrency issue; the resulting harness replication materially strengthened the empirical case of the paper.

% --- Bibliography ----------------------------------------------------------
\bibliographystyle{plainnat}
\bibliography{references}

% =========================================================================
\appendix

\section{Per-seed summaries}
\label{app:perseed}

\cref{tab:seed-concurrent,tab:seed-sequential} expand the aggregate statistics to the per-seed level. With only five seeds, the reader should be able to see every point that enters the non-parametric tests.

\begin{table}[H]
\centering
\small
\caption{Per-seed results under the corrected concurrent harness. Fast = avg.\ request latency $<1000$\,ms. Post-hit consistency is the fraction of later trials that remain fast after the first fast trial.}
\label{tab:seed-concurrent}
\begin{tabular}{llcccc}
\toprule
Optimizer & Seed & Fast\,/\,15 & Post-hit & First fast trial & Best latency (ms) \\
\midrule
Random    & 42  & 9  & $0.727$ & 4 & $464.85$ \\
Random    & 142 & 8  & $0.636$ & 4 & $461.42$ \\
Random    & 242 & 3  & $0.167$ & 3 & $486.38$ \\
Random    & 342 & 9  & $0.667$ & 3 & $465.88$ \\
Random    & 442 & 8  & $0.500$ & 1 & $474.05$ \\
\midrule
\sloguard{} & 42  & 9  & $1.000$ & 7 & $467.54$ \\
\sloguard{} & 142 & 9  & $1.000$ & 7 & $468.04$ \\
\sloguard{} & 242 & 11 & $0.833$ & 3 & $462.96$ \\
\sloguard{} & 342 & 11 & $0.833$ & 3 & $466.19$ \\
\sloguard{} & 442 & 11 & $0.714$ & 1 & $463.73$ \\
\bottomrule
\end{tabular}
\end{table}

\begin{table}[H]
\centering
\small
\caption{Per-seed results under the earlier sequential harness. Absolute latencies differ from the concurrent harness because sequential dispatch under-stresses the system; the consistency ranking is preserved.}
\label{tab:seed-sequential}
\begin{tabular}{llcccc}
\toprule
Optimizer & Seed & Fast\,/\,15 & Post-hit & First fast trial & Best latency (ms) \\
\midrule
Random    & 42  & 9  & $0.727$ & 4 & $429.40$ \\
Random    & 142 & 8  & $0.636$ & 4 & $429.14$ \\
Random    & 242 & 3  & $0.167$ & 3 & $431.59$ \\
Random    & 342 & 9  & $0.667$ & 3 & $432.80$ \\
Random    & 442 & 8  & $0.500$ & 1 & $432.61$ \\
\midrule
\sloguard{} & 42  & 11 & $1.000$ & 5 & $432.00$ \\
\sloguard{} & 142 & 9  & $1.000$ & 7 & $430.01$ \\
\sloguard{} & 242 & 11 & $0.833$ & 3 & $431.41$ \\
\sloguard{} & 342 & 11 & $0.833$ & 3 & $434.55$ \\
\sloguard{} & 442 & 11 & $0.714$ & 1 & $429.86$ \\
\bottomrule
\end{tabular}
\end{table}

\section{Metric definitions}
\label{app:metrics}

\paragraph{Fast-cluster trial.} A feasible trial is counted as fast-cluster when its average request latency is below $1000$\,ms. The threshold is chosen to cleanly separate the two visibly distinct latency bands in \cref{fig:bimodal} rather than to maximize the method's apparent advantage.

\paragraph{Post-hit consistency.} Let $t^\star$ be the first trial whose average request latency is below $1000$\,ms. Post-hit consistency is
\[
\mathrm{PHC} \;=\; \frac{\#\{t>t^\star:\text{trial }t\text{ is fast-cluster}\}}{\#\{t>t^\star\}}.
\]
It answers: \emph{once the fast regime is discovered, how often does the optimizer continue to spend later trials there?}

\paragraph{Best latency.} $\min_t\text{avg.~latency}(t)$ over feasible trials for a given $(\text{optimizer},\text{seed})$ pair. Reported because it is standard and easy to compare, but deliberately not the headline metric because the mean difference between optimizers is not statistically persuasive.

\paragraph{Directionality of tests.} The directional hypothesis is that a crash-aware two-phase optimizer should spend more of a fixed budget in the right regime than random search, not less; hence one-sided tests on fast-cluster count and post-hit consistency. No corresponding directional claim is made on best latency, so the best-latency test is two-sided.

\end{document}